\documentclass[sigconf]{acmart}

\AtBeginDocument{%
  }

\setcopyright{acmlicensed}
\copyrightyear{2018}
\acmYear{2018}
\acmDOI{XXXXXXX.XXXXXXX}

\acmConference[Conference acronym 'XX]{Make sure to enter the correct
  conference title from your rights confirmation emai}{June 03--05,
  2018}{Woodstock, NY}
\acmISBN{978-1-4503-XXXX-X/18/06}

\usepackage{multirow}
\usepackage{subfig}

\begin{document}

\title{Multi-Margin Cosine Loss: Proposal and Application in Recommender Systems}


\author{Makbule Gulcin Ozsoy}
\affiliation{%
  \city{London}
  \country{UK}}
\email{makbulegulcin@gmail.com}

\renewcommand{\shortauthors}{Ozsoy et al.}

\begin{abstract}
Recommender systems guide users through vast amounts of information by suggesting items based on their predicted preferences. Collaborative filtering-based deep learning techniques have regained popularity due to their straightforward nature, relying only on user-item interactions. Typically, these systems consist of three main components: an interaction module, a loss function, and a negative sampling strategy. Initially, researchers focused on enhancing performance by developing complex interaction modules. However, there has been a recent shift toward refining loss functions and negative sampling strategies. This shift has led to an increased interest in contrastive learning, which pulls similar pairs closer while pushing dissimilar ones apart. Contrastive learning may bring challenges like high memory demands and under-utilization of some negative samples. The proposed Multi-Margin Cosine Loss (MMCL) addresses these challenges by introducing multiple margins and varying weights for negative samples. It efficiently utilizes not only the hardest negatives but also other non-trivial negatives, offers a simpler yet effective loss function that outperforms more complex methods, especially when resources are limited. Experiments on two well-known datasets demonstrated that MMCL achieved up to a 20\% performance improvement compared to a baseline loss function when fewer number of negative samples are used.
\end{abstract}

\ccsdesc[500]{Information systems~Recommender systems}
\ccsdesc[500]{Computing methodologies~Neural networks}

\keywords{Recommender systems, Loss function}



\maketitle

\section{Introduction} \label{intro}



\begin{figure}
    \centering
     \includegraphics[width=1.0\linewidth]{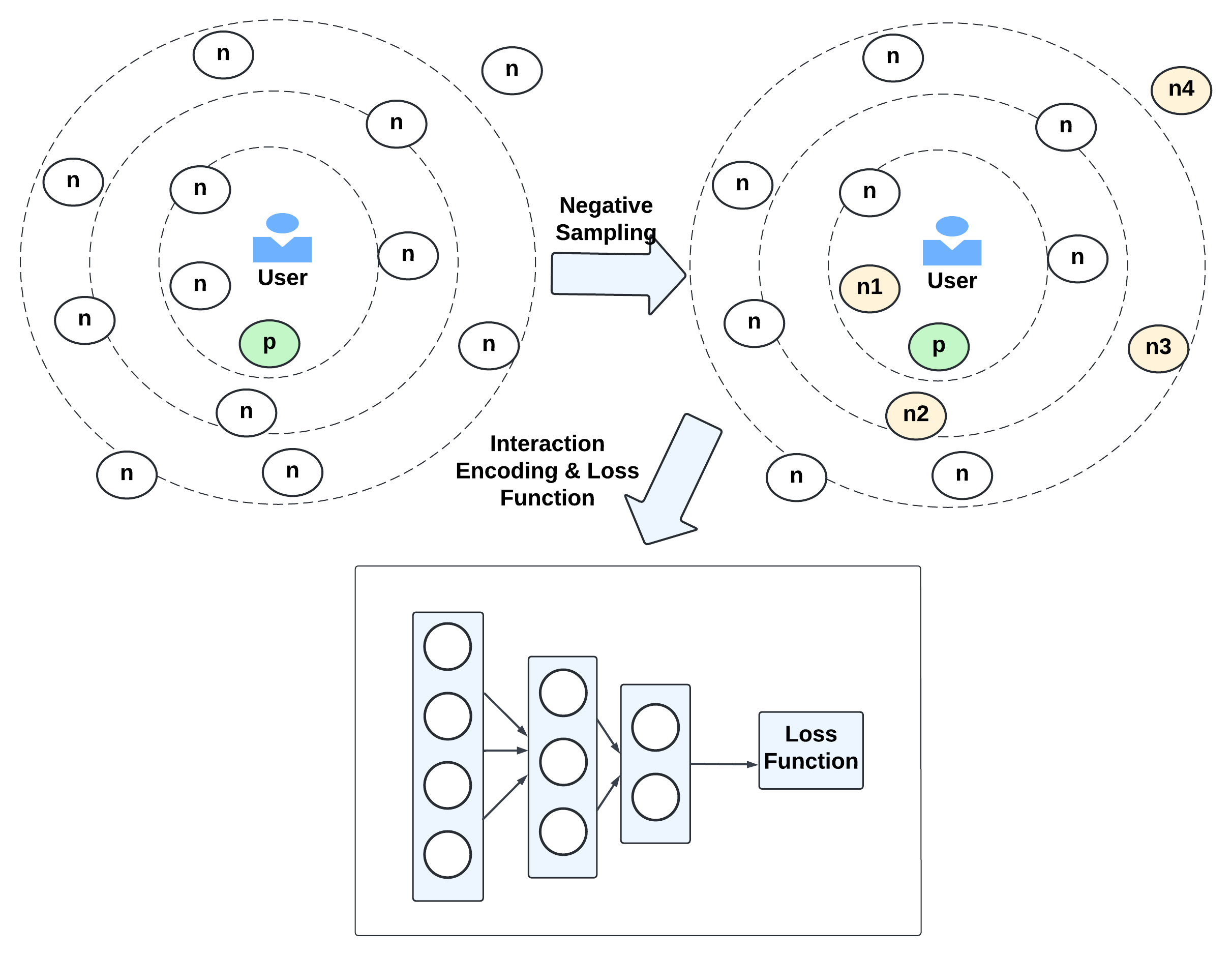}
         \caption{
    Three major components for a deep-learning based CF model: (i) interaction encoder, (ii) loss function and (iii) negative sampling. 
    A user interacts with an item  $p$ (e.g., purchases a product). From the remaining negative items $n$, four negatives, $n_1$-$n_4$, are sampled. The sampled negative items, positive item and user representations are fed into the interaction encoding and loss function modules for computation.}
    \Description{Three major components for a deep-learning based CF model: (i) interaction module/encoder, (ii) loss function and (iii) negative sampling strategy.}
    \label{fig:stages}
\end{figure}

Numerous web applications like YouTube, Amazon, Spotify use recommendation systems (RS) suggest items such as videos, products, and music tracks by estimating user preferences.
RS utilize various methods, from traditional methods like collaborative filtering (CF), content-based filtering, matrix factorization \cite{li2015rank, HeLLSC16} to
more recent techniques like deep learning, reinforcement learning, language models \cite{ozsoy2016word, he2017neural, musto2018deep, zheng2018drn, shi2021dares, sun2019bert4rec, fan2023recommender}. 
Recently, CF-based deep learning techniques have regained popularity due to their simplicity, efficacy, and reliance solely on user-item interactions. 
These methods are composed of three main components: interaction encoder, loss function and negative sampling strategy, 
Figure \ref{fig:stages} illustrates these steps: A user interacts with an item $p$ (e.g., purchases a product), and four negative items, $n_1$-$n_4$, are sampled. The sampled negative items, positive item and user representations are then used by the interaction encoding and loss function modules for the computations and making recommendations.

SimpleX \cite{mao2021simplex} demonstrated that a simple interaction encoding model with a well-designed loss function can outperform more complex techniques, shifting focus in RS towards loss functions and negative sampling strategies.
Recent studies have analyzed the impact of loss functions on recommendation performance, 
and developed new ones to address challenges like popularity bias and robustness \cite{wu2022effectiveness, li2023revisiting, park2023toward, zhang2022incorporating, sun2023neighborhood, wu2023bsl, zhang2024empowering}. 
These efforts increased the interest in contrastive learning within RS domain.

Contrastive learning models aim to learn representations by bringing similar pairs closer while separating dissimilar pairs. 
For an efficient contrastive learning system, three key components have been identified \cite{weng2021contrastive}: heavy data augmentation, large batch size and hard-negative sampling. While using larger batch sizes and more negative samples introduce more diverse and `hard'-negative samples to the system, they  require more memory as a side-effect. 
A common approach is to use a margin parameter to distinguish `hard'-negatives from the rest. However, this approach sets only a single threshold, failing to fully utilize all available negative samples efficiently.

In this work, we hypothesise that using all non-trivial negative instances, not just the hardest, allows the model to generalize more efficiently, even with fewer negative samples or smaller batch sizes. 
Assigning different weights to varying levels of "hardness" in negative instances—such as giving higher weight to the hardest negatives and relatively lower weights to easier ones—can enhance performance.
For this purpose, we propose Multi-Margin Cosine Loss (MMCL), a simple yet efficient loss function, that uses multiple margins and assigns different weights to different levels of negative instances.
The main contributions of this work are:
\begin{itemize}
    \item We proposed Multi-Margin Cosine Loss (MMCL), which introduces multiple margins and weights for negative instances. It allows consideration of not only the hardest negatives but also 'semi-hard' and 'semi-easy' negatives. 
    The ability of assigning different weights enables tuning the importance of each level of negative samples.
    \item We conducted experiments on two well-known recommender system datasets, comparing MMCL's performance to state-of-the-art loss functions.
    \item MMCL performs comparably to state-of-the-art contrastive loss functions with a large negative sample size (e.g., 800) and achieves up to a 20\% performance improvement over baseline loss functions with smaller sample sizes (e.g., 10 or 100).
\end{itemize}

The paper is structured as follows: Section \ref{relWork} provides background on recommender systems, contrastive learning and loss functions. Section \ref{proposedWork} presents the proposed Multi-Margin Cosine Loss (MMCL). Section \ref{eval} details the experimental setup and evaluation results. Section \ref{conclusion} discusses the conclusion and future directions.

\section{Related Work}\label{relWork}

In this section, background information on recommender systems (RS), 
overview of loss functions used in RS and details on contrastive loss functions are presented. 

\begin{figure*}
    \centering
    \subfloat[]{
        \includegraphics[width=.37\linewidth]{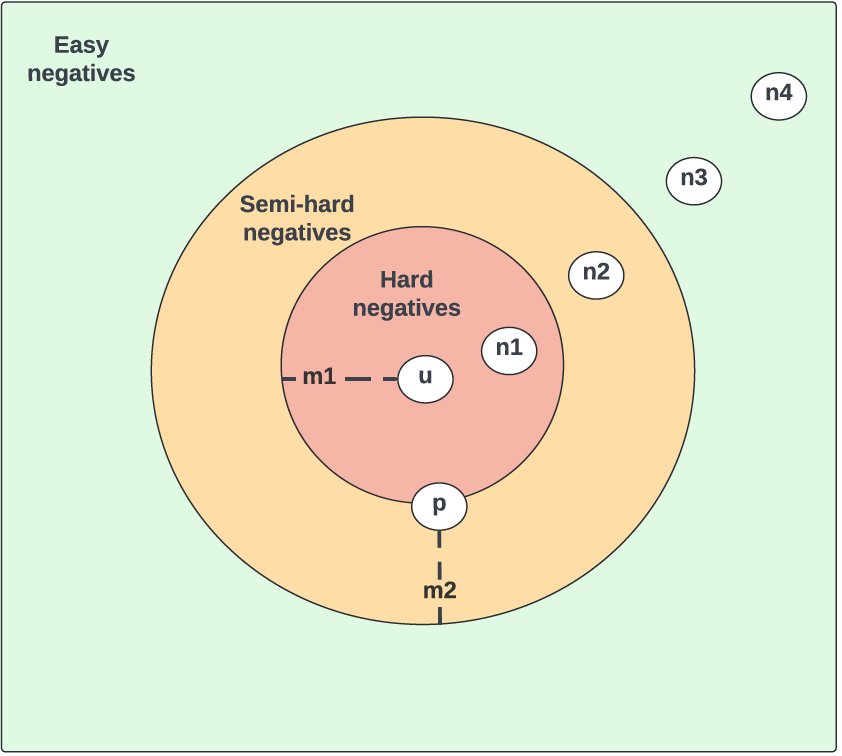}
        \label{subfig:ccl_triplet}
    } \quad \quad \quad \quad
    \subfloat[]{
        \includegraphics[width=.37\linewidth]{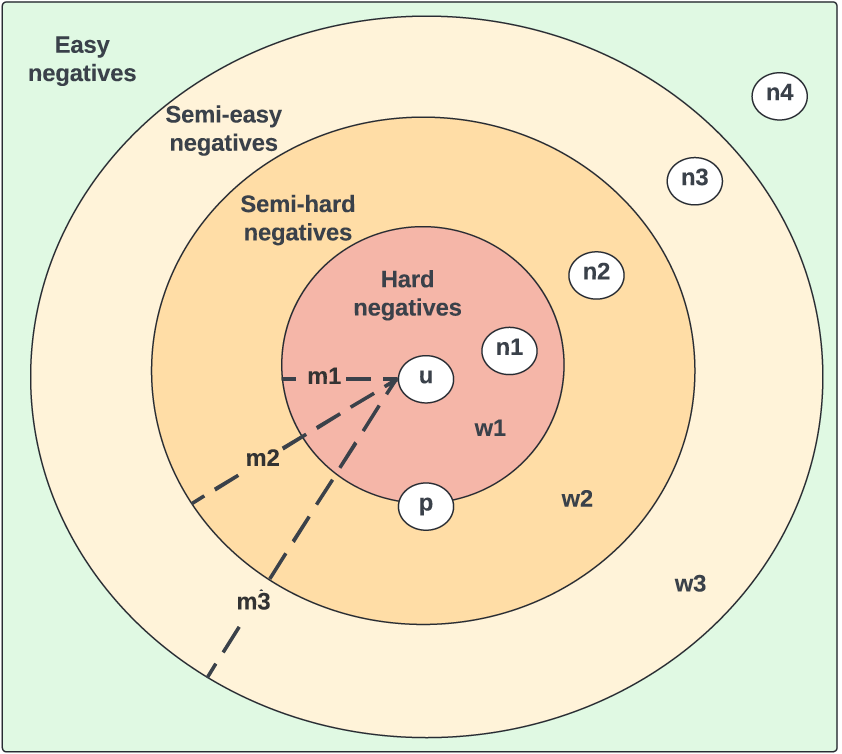}
        \label{subfig:mml}
    }
    \caption{(a) Margin $m$ defines a radius, which makes the model to pay attention to harder-negatives. In Contrastive Loss, Eq.\ref{eq:contrastiveLoss}, the radius is relative to user representation ($m1$). In Triplet Loss, Eq. \ref{eq:tripletLoss} it is relative to positive item ($m2$). (b) Proposed Multi-Margin Cosine Loss (MMCL), Eq. \ref{eq:multiMarginLoss}, filters not only the hardest-negatives but also other non-trivial negatives, using multiple margins, $m1,m2,m3$, and assigns different weights to each level, $w1,w2,w3$.}
    \Description{Margin $m$ defines a radius, which makes the model to pay attention to harder-negatives.}
    \label{fig:loss_functions}
\end{figure*}

\subsection{Collaborative Filtering-based Recommender Models}
Recommender systems (RS) estimate users' preferences and recommend items based on historical user-item interaction. 
Recently, CF-based deep learning techniques have regained popularity due to their simplicity, efficacy, and reliance solely on user-item interactions.  
These systems 
consist of three main components (Figure \ref{fig:stages}):
\textit{\textbf{(i) Interaction encoder}} learns embeddings of each user and item. It has been primary focus in the field for many years and various techniques have been employed, such as vector embeddings, multi-layer perceptrons, auto-encoders, attention networks, transformers, graph neural networks, reinforcement learning techniques \cite{ozsoy2016word, he2017neural, musto2018deep, sun2019bert4rec, chen2020revisiting, zheng2018drn, ozsoy2020mp4rec, shi2021dares}. 
Although these methods improve performance, their complexity has increased. Efficient models are in higher demand for practical applications.
\textit{\textbf{(ii)  Loss function}}: Common loss functions in RS include Binary Cross-Entropy (BCE) \cite{he2017neural}, Bayesian Personalized Ranking (BPR) \cite{rendle2009bpr} or mean square error (MSE) \cite{chen2020efficient}. 
Inspired by advancements in computer vision and natural language processing, new loss functions utilizing both positive and negative instances and contrastive learning approaches receive increased focus in RS applications.
\textit{\textbf{(iii) Negative sampling}} samples a subset of negative instances from a larger set. In RS, there are vast amount of negative or unobserved instances compared to positive ones and to improve training efficiency negative sampling techniques are applied. Various strategies exist in the literature, such as random, popularity-based and GAN-based sampling \cite{ding2019reinforced, barkan2016item2vec, lian2020personalized, zhang2013optimizing, jin2020sampling, yang2024does}. 

\subsection{Overview of Loss Functions used in Recommender Systems} \label{loss_rs}

There are three common types of loss functions in RS: 
\textit{\textbf{(i) Point-wise loss}} measures the distance between predicted and expected labels, using each instance independently, i.e., either a positive or a negative instance is used at a time. Examples in RS include Binary Cross Entropy~(BCE) \cite{he2017neural}, Mixed-centric~(MCL) \cite{gao2022mcl} and User Interest Boundary~(UIB) \cite{zhuo2022learning} losses. 
\textit{\textbf{(ii) Pair-wise loss}} compares pair of items to decide the optimal ordering. 
Examples used in RS include Bayesian Personalized Ranking (BPR) \cite{rendle2009bpr} and Collaborative Metric Learning (CML) \cite{hsieh2017collaborative} losses
\textit{\textbf{(iii) Set-wise loss}} generalizes pairwise loss functions by considering multiple negative items for a single positive item. Examples used in RS include  Cosine Contrastive Loss (CCL) \cite{mao2021simplex}, Sampled Softmax (SSM) \cite{wu2022effectiveness}, Bias-aware Contrastive (BC) \cite{zhang2022incorporating} losses. 

Another direction in RS involves using automated machine learning (AutoML) techniques to explore loss functions, such as AutoLoss\cite{zhao2021autoloss}, AutoLossGen \cite{li2022autolossgen}, which combine or generate new loss functions automatically. 
Well-designed loss functions and AutoML techniques complement each other and enhance RS performance.
For instance, SimpleX \cite{mao2021simplex} showed that a well-designed loss function enables a simple model to outperform more complex ones, sparking interest in loss functions for RS. 
Recent studies have analyzed the impact of loss functions on RS performance and developed new ones to address challenges, like popularity bias \cite{wu2022effectiveness, li2023revisiting, park2023toward, wu2023bsl, zhang2022incorporating, sun2023neighborhood, zhang2024empowering},  highlighting contrastive learning as a significant trend. 

\subsection{Contrastive Learning} \label{contrastive_learning}
In machine-learning, 
contrastive learning gained attention attention for its better performance, as it learns representations by bringing similar pairs closer and separating dissimilar ones.

\textit{\textbf{Scientific Notation}}: Given a set of users, $U$, and items, $I$, user-item interaction can be represented as $R \in \{0,1\} ^{|U|x|I|}$, where $r_{ui}=1$ indicates positive interactions and $r_{ui}=0$ indicate negative interactions. 
Positive interactions form the set of positive instances: $I_{u}^+=\{i\in I | r_{ui}=1\}$ and negative interactions form the set of negative instances: $I_{u}^-=\{i\in I | r_{ui}=0\}$. 
Various loss functions, $f$, are used for training, utilizing predicted relevance $s(u,i)$, predicted distance $d(u,i)$ and actual label $l(u,i)$ of the user-item interactions. Some loss functions also use margin $m$ and weight $w$ parameters.
Common contrastive loss functions are summarized below: \\
\textit{\textbf{Contrastive loss \cite{chopra2005learning}}} is one of the earliest contrastive loss functions. 
For a user $u$ and item $i$ (i.e., $i\in I+$ or $i\in I-$), the goal is to make positive items closer and the negative items further away from the user. Its equation is: 

\begin{equation} \label{eq:contrastiveLoss}
\begin{split}
L_{Contr.} = \dfrac{1}{2} \sum_{{x}\in{I}} Y_i d(u,i)^2 {\ +} 
    (1-Y_i) max(0, m-d(u,i))^2
\end{split}
\end{equation}
In the equation, $Y_i$ function returns 1 if item is positive (i.e., $i \in I+$) and 0 otherwise. Margin $m$ defines a radius making the model to pay attention to hard-negatives, which are negative instances which has the lowest distance (i.e., most similar) to the user representation. 
Figure \ref{subfig:ccl_triplet} illustrates this concept, where margin $m1$ filters negative samples. As the result, only $n1$, which is in the hard-negatives area, and positive item $p$ are used for the computations.

Following a similar concept, Cosine Contrastive Loss (CCL), proposed by SimpleX \cite{mao2021simplex}, is one of the first application of contrastive loss in RS. CCL utilizes a set of negative items and cosine similarity instead of distance. Its equation is:
\begin{equation} \label{eq:ccl}
\begin{split}
L_{CCL} = \sum_{{i}\in{I+}, {N}\subset{I-}} 1 - s(u,i) {\ +} 
    \dfrac{w}{N} \sum_{{j}\in{N}} max(0, s(u,j)-m)
\end{split}
\end{equation}    
\textit{\textbf{Triplet loss \cite{schroff2015facenet}}} is another commonly used contrastive loss function, originally proposed for face recognition in FaceNet \cite{schroff2015facenet}. Using triplets of user $u$, positive item $i \in I+$ and negative item $j \in I-$, triplet loss aims to minimize distance between the user and the positive item while maximizing the distance between the user and the negative item. Its equation: 
\begin{equation} \label{eq:tripletLoss}
  L_{Triplet} = \sum_{i\in I+, j\in I-}  max(0, d(u,i)^2 - d(u,j)^2 + m)
\end{equation}
In the equation, margin $m$ defines a radius: 
(i) If the distance of negative item to the user is less than the distance of the positive item, $d(u,n) < d(u,p)$, it is considered a hard negative. (ii) If the distance of negative item is between the distance of the positive item and the sum of the positive item distance and margin, $d(u,p) < d(u,n) < d(u,p)+m$, it is a semi-hard negative. 
Both hard and  semi-hard negatives are included in the computations. 
In Figure \ref{subfig:ccl_triplet}, the margin $m2$ is used for filtering negative samples. As the result, $n1$ and $n2$, which are in the radius of positive item $p$, are used for the computations.\\
\textit{\textbf{InfoNCE loss \cite{oord2018representation}}} builds on Noise Contrastive Estimation (NCE) \cite{gutmann2010noise}, which distinguishes the target data from noise. It ranks the positive samples higher. 
Its equation is: 
\begin{equation} \label{eq:infoNCELoss}
    L_{InfoNCE} = - \sum_{i \in I+} log \dfrac{exp(s(u,i))}{exp(s(u,i)) + \sum_{j\in N}{exp(s(u,j))}}
\end{equation}
AdvInfoNCE\cite{zhang2024empowering} enhances InfoNCE loss by adaptively exploring and assigning a difficulty level to each negative instance in an adversarial manner.\\
\textit{\textbf{Bilateral SoftMax Loss \cite{wu2023bsl}}} is a recent addition to RS loss functions. It extends softmax loss (SL) by applying normalization techniques to both positive and negative instances, enhancing robustness to noisy data. 
Its equation is:
\begin{equation} \label{eq:BSL}
\begin{split} 
    L_{BSL} = -\tau_1 log \mathbb{E}_{i\sim P+} \dfrac{exp(s(u,i))}{\tau_1} {\ +}  
        \tau_2 log \mathbb{E}_{j\sim P-} \dfrac{exp(s(u,j))}{\tau_2}
\end{split} 
\end{equation}
In the equation, $P$ is the distribution of instances, $\mathbb{E}$ is the expectation, and $\tau$ are the temperatures of components.


\section{Proposed Loss Function}\label{proposedWork}
In this section, discuss the generalization of contrastive loss functions, explain the rationale behind the proposed Multi-Margin Cosine Loss (MMCL) function and provide its details. 

\subsection{Generalized Contrastive Loss}\label{generalizedCL}
From the loss functions summarized in the related work section, we can generalize the contrastive loss as follows:
\begin{equation} \label{eq:generalizedLoss}
\begin{split}
    L_{\substack{Gen\\Contr}} = \sum_{i\in I+, N\subset I-} w_p f(u,i) {\ +} 
        \dfrac{w_n}{N} \sum_{j \in N} max(0, f(u,i,j, m))
\end{split}
\end{equation}
In the equation: 
(i) We used a set-wise loss, computing the loss over a positive instance and a set of N negative samples. Using only one negative sample reduces it to pair-wise loss.
(ii) We assigned distinct weights to positive and negative instances, $w_p$ and $w_n$. Previously, in Contrastive Loss or CCL, $w_p$ is set to 1.0, and in Triplet Loss to 0.0.
(iii) We employed a generic function $f(u,i,j,m)$ to represent various combinations of margin $m$, positive and negative instances, $i$ and $j$, respectively.
The function can vary: InfoNCE does not use a margin; $s(u,j)$.
CCL uses similarity between the user and the negative instance and margin; $s(u,j)-m$.
Contrastive Loss uses a distance function and margin; $m-d(u,j)$. 
Triplet Loss uses positive and negative items and margin; $d(u,i)-d(u,j)+m$, where margin is relative to the positive item. 

\subsection{Motivation for Multi-Margin Cosine Loss}\label{motivationMMCL}

Effective contrastive learning relies on three key components \cite{weng2021contrastive}: 
\textit{(i) Heavy Data Augmentation} techniques generate noisy versions of positive instances to enhance feature generalization; 
\textit{(ii) Large Batch Size} provides a diverse and challenging set of negative samples, although it can be memory-intensive;
\textit{(iii) Hard-Negative Mining} identifies negative samples similar to the anchor (user) but which are negative, i.e., user did not like that item.
A solution for incorporating hard negatives is to use large batch sizes or more negative samples.  
However, this approach suffers from higher memory usage as a side effect. 
A more efficient solution is to use a margin $m$ in computations to distinguish hard negatives efficiently.
In contrastive learning, it is important to effectively utilize negative samples while prioritizing the more challenging ones.
Instead of relying on larger batch sizes for introducing more hard negatives, we propose using not only the most challenging but all non-trivial instances, 
and assigning weights based on their "hardness".
This approach allows effective learning with fewer samples or smaller batch sizes, from a diverse set of negative instances.

Training an RS model with contrastive learning helps distinguish similar and dissimilar pairs and learn representations that capture meaningful relationships between users and items. 
This improves RS performance by generalizing effectively with limited interaction data, understanding a wider range of item similarities and dissimilarities, and ensuring that the negative pairs are diverse and unbiased. 
We propose Multi-Margin Cosine Loss (MMCL) to address these needs, which enhances contrastive learning by considering multiple margins and weights for improved representation and performance.

\subsection{Multi-Margin Cosine Loss}\label{mmcl}
The proposed Multi-Margin Cosine Loss (MMCL) introduces multiple margins and weights for negative instances, considering hardest-negatives other non-trivial negatives; i.e., 'semi-hard', 'semi-easy'-negatives. 
Weights enable tuning the importance of each level of negatives.
Using a similar approach to CCL \cite{mao2021simplex}, we extended the generalized contrastive loss.
The MMCL equation is as follows:

\begin{equation} \label{eq:multiMarginLoss}
\begin{split}
    L_{MMCL} = \sum_{i\in I+, N\subset I-} w_p (1-s(u,i)) {\ +}  \\
        \dfrac{1}{N} \sum_{j \in N} \sum_{k\in |M|} w_k max(0, s(u,j)-m_k)
\end{split}
\end{equation}
In the equation, $|M|$ is the size of the defined margins. Each margin $m_k \in M$ is assigned a different weight, $w_k$. 
Similar to CCL, (i) similarity is used for computations: $f(u, i) = 1- s(u, i)$ and $f(u,i,j, m_k) = s(u,j)-m_k$ and (ii) margins are set relative to the user representation.
Figure \ref{subfig:mmcl} illustrates MMCL: 
It uses multiple margins, $m1,m2,m3$ $m1$, and assigns different weights $w1, w2, w3$. Consequently, not only hardest-negatives, e.g., $n1$, but other non-trivial negatives, e.g., $n2$ and $n3$, are selected and used in the computations.

In this section, we introduce the Multi-Margin Loss (MML). We start by discussing the generalization of contrastive loss functions, then we explain the rationale behind the MML function. Finally, we provide a detailed description of the MML function itself.

 \begin{table}
\caption{The statistics of the datasets}\label{table:dataset}
\begin{center}
\begin{tabular}{ccc} 
 \toprule
 {} & \textbf{Yelp} & \textbf{Gowalla} \\
 \midrule
 \textbf{\#Users} & {31,668} & {29,858}\\
 \textbf{\#Items} & {38,048} & {40,981}\\
 \textbf{\#Interactions} & {1,561,406} & {1,027,370}\\ 
 \textbf{\#Train} & {1,237,259} & {810,128}\\
 \textbf{\#Test}  & {324,147} & {217,242}\\
 \textbf{Density} & {0.00130} & {0.00084}\\
\bottomrule
 \end{tabular}
\end{center}
\end{table}

\begin{table*}
\caption{Performance comparison of MMCL and CCL on SimpleX and MF architectures 
(\#Negative samples = 100).  
Among the all settings, MMCL consistently delivers the best results on the Yelp and Gowalla datasets.
}\label{table:compare_earlystop}
\begin{center}
\begin{tabular}{ccccccc} 
 \toprule
 
 \multirow{2}{*}{Model} & \multirow{2}{*}{Loss} & \multirow{2}{*}{Early-stop?} & \multicolumn{2}{c}{Yelp18} & \multicolumn{2}{c}{Gowalla} \\ \cline{4-5} \cline{6-7}  
{}& {} & {} & Recall@20 & Ndcg@20 & Recall@20 & Ndcg@20\\
  \midrule

  \multirow{4}{*}{SimpleX} & CCL (2023) & Y  & 0.0634 & 0.0517 &  0.1622 & 0.1283\\ 
{}&CCL (2023) & N & 0.0679 & 0.0555 & 0.1663 & 0.1319\\
{}&MMCL (2024) - \textit{Proposed} & Y & 0.0662 & 0.0539 & 0.1723 & 0.1358\\ 
{}&MMCL (2024) - \textit{Proposed} & N & \textbf{0.0706} & \textbf{0.0578} &  \textbf{0.1787} & \textbf{0.1412}\\ 
\midrule
\multirow{4}{*}{MF} & CCL (2023) & Y & 0.0601 &0.0490 & 0.1632 & 0.1279\\ 
{}&CCL (2023) & N  & 0.0673 & 0.0551 & 0.1783 & 0.1426\\
{}&MMCL (2024) - \textit{Proposed} & Y & 0.0674 & 0.0550 &  0.1756 & 0.1370 \\ 
{}&MMCL (2024) - \textit{Proposed} & N & \textbf{0.0702} & \textbf{0.0580} & \textbf{0.1816} & \textbf{0.1435} \\
            
\bottomrule
\end{tabular}
\end{center}
\end{table*}

\begin{table*}
\caption{Training-Runtime Comparison on MF model (Without early-stopping, 100 epochs). 
MMCL increases training time per epoch by at most 2 seconds on average.
}\label{table:compare_efficiency}
\begin{center}
\begin{tabular}{cccccccc} 
 \toprule
\multirow{2}{*}{\#Negative samples} &  \multirow{2}{*}{Loss} & 
        \multicolumn{3}{c}{Yelp18} & \multicolumn{3}{c}{Gowalla} \\ \cline{3-5} \cline{6-8}  
{} & {} & Time per epoch & Ratio & Total time & Time per epoch & Ratio & Total time\\
\midrule 
\multirow{3}{*}{800}  & CCL (2021) & 77.5 secs & - & $\sim$129 mins & 66 secs  & - & 110 mins \\
{} & MMCL (2024) - \textit{Proposed} & 78.5 secs & +1.3\% & $\sim$130 mins  & 64 secs  & -3.0\% & $\sim$107 mins\\
\midrule
\multirow{3}{*}{100}  & CCL (2021) & 35 secs & - & $\sim$58 mins & 37 secs & - & $\sim$62 mins \\
{} & MMCL (2024) - \textit{Proposed} & 35 secs & 0.0\%  & $\sim$58 mins &  37.5 secs &+1.4\% & $\sim$63 mins\\
\midrule
\multirow{3}{*}{10}  & CCL (2021) & 31 secs & - & $\sim$51 mins & 32 secs & - & $\sim$53 mins\\
{} & MMCL (2024) - \textit{Proposed} & 33 secs & +6.5\% & 55 mins & 33.5 secs & +4.7\% & $\sim$56 mins \\

\bottomrule
 \end{tabular}
\end{center}
\end{table*}

\section{Evaluation Results}\label{eval}

\subsection{Evaluation Setup}\label{dataset_evalMetrics_expParameters}
For comparisons, 
six loss functions are used: (i) four classical loss functions: BPR, Pairwise Hinge, Softmax Cross-Entropy, and Mean Square Error; (ii) two recently proposed contrastive loss functions: CCL \cite{mao2021simplex} and BSL \cite{wu2023bsl}. 



We use two commonly used datasets in RS, particularly by recent neural-network-based CF models: (i) Yelp2018 \cite{yelp2018dataset}\footnote{Yelp2018: \url{https://huggingface.co/datasets/reczoo/Yelp18_m1}} and (ii) Gowalla \cite{cho2011friendship}\footnote{Gowalla: \url{https://huggingface.co/datasets/reczoo/Gowalla_m1}}. 
These datasets are pre-split, pre-processed, and publicly available and their statistics are summarized in Table~\ref{table:dataset}. 
We use the datasets as is, without any additional preprocessing or filtering based on popularity or cold-start. 
All not interacted items are considered candidates, and average results across all users are reported.
Performance is evaluated using Recall and NDCG metrics @20, as used by comparison partners, CCL and BSL.

Computations were performed on a P100 GPU with a 30-hour weekly limit, focusing on performance enhancement within this constraint. 
The MMCL implementation expanded on SimpleX's code and is shared online\footnote{MMCL (Proposed) code: \url{https://github.com/mgulcin/mmcl}}. 
Experiments using CCL employed SimpleX code\cite{mao2021simplex}\footnote{SimpleX (CCL) \cite{mao2021simplex} code: \url{https://reczoo.github.io/SimpleX}} and BSL experiments used BSL code\cite{wu2023bsl}\footnote{BSL \cite{wu2023bsl} code: \url{https://github.com/junkangwu/BSL}}. All experiments were rerun, and results are presented below.
Parameters followed the SimpleX paper: Adam optimizer with $L_2$ regularization, a 1e-4 learning rate, a 64 embedding dimension, and 100 epochs. 
The batch size was 512 for Yelp and 256 for Gowalla, with regularizers of 1e-09 and 1e-08, respectively. 
Results with 100 negative samples are presented, unless otherwise specified.
The MMCL parameters were determined via grid search.
For Yelp, $w_p$:$w_n$ ranged from [1:150] to [1:450] in increments of 50, and for Gowalla from [1:500] to [1:1000] in increments of 100. 
Margin values started at [0.9] and expanded to [0.5, 0.6, 0.7, 0.8, 0.9]. Margin weights were adjusted with priority given to the highest margin and reductions made proportionally. 
As a result, for Yelp, margins set to [0.6, 0.7, 0.8, 0.9] with weights [0.05, 0.15, 0.2, 0.6] and $w_p$:$w_n$ set to [1:400]. For Gowalla, margins set to [0.7, 0.8, 0.9] with weights of [0.10, 0.15, 0.75] and $w_p$:$w_n$ set to [1:700].
Baseline loss functions CCL and BSL were rerun using their publicly available code and configurations.

\begin{table*}
\caption{Performance of MF model with different loss functions (\#Negative samples = 100). 
MMCL performs equally well in recall and Ndcg on the Yelp dataset and achieves the best results in Recall on the Gowalla dataset.
}\label{table:compare_loss_functions}
\begin{center}
\begin{tabular}{ccccc} 
 \toprule
 \multirow{2}{*}{Loss} & 
        \multicolumn{2}{c}{Yelp18} & \multicolumn{2}{c}{Gowalla} \\ \cline{2-3} \cline{4-5}  
{} & Recall@20 & Ndcg@20 & Recall@20 & Ndcg@20\\
\midrule 
BPR Loss & 0.0272 & 0.0216 &  0.0708 & 0.0509\\
Pairwise Hinge Loss & 0.0340 & 0.0269 &   0.0758 & 0.0541\\
Softmax Cross-Entropy & 0.0309 & 0.0246 &   0.0734 & 0.0543\\
Mean Square Error & 0.0609 & 0.0502 &  0.1391 & 0.1171\\
        
\midrule 
CCL (2021) & 0.0673  & 0.0551 &  0.1783 & 0.1426 \\
BSL (2023) & \textbf{0.0703} & \textbf{0.0580} &   0.1804 & \textbf{0.1521}  \\
MMCL (2024) - \textit{Proposed} & 0.0702 & \textbf{0.0580} & \textbf{0.1816} & 0.1435 \\
\bottomrule
 \end{tabular}
\end{center}
\end{table*}

\begin{figure*} [!t]%
    \centering
    \subfloat[\centering Yelp dataset]{{\includegraphics[width=0.4\linewidth]{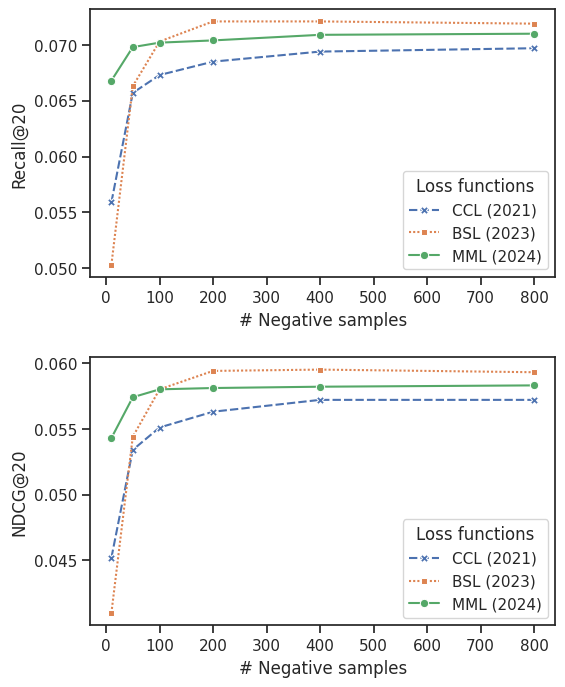} }}%
    \qquad %
    \subfloat[\centering Gowalla dataset]{{\includegraphics[width=0.39\linewidth]{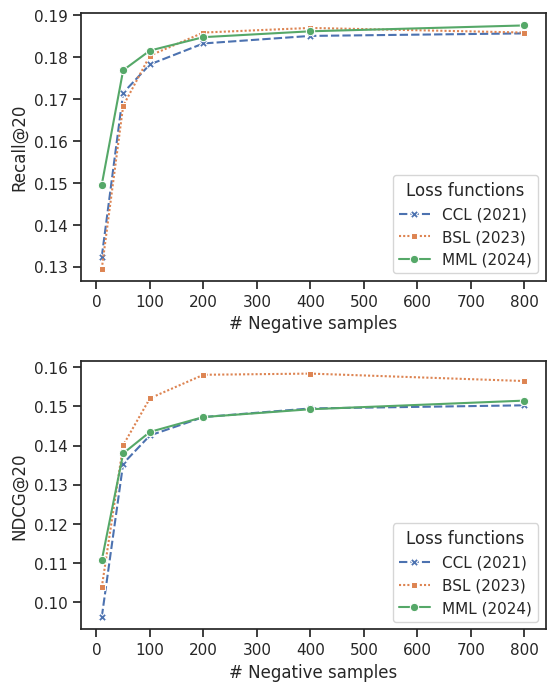} }}%
    \caption{Performance of MF model under recent loss functions. 
    The proposed MMCL outperforms baseline loss functions when the number of negative samples is 100 or fewer, suggesting it is more efficient in resource-constrained environments.
    }%
    \label{fig:compare_number_of_negatives}%
\end{figure*}

\subsection{Evaluation Results}\label{evalResults}
\textit{\textbf{Impact of model architecture and early-stopping}}
SimpleX demonstrated that their CCL loss function, combined with simpler neural network architectures, like SimpleX and Matrix Factorization (MF), outperforms more complex models.
They used early-stopping, which halts training when validation performance no longer improves. 
Table~\ref{table:compare_earlystop} compares the performance of CCL and MMCL on SimpleX and MF architectures, with and without early-stopping. With early-stopping, SimpleX architecture concluded after 33 epochs with CCL and 19 epochs with MMCL, while the MF architecture ended after 19 epochs for both loss functions. 
The table shows that omitting early-stopping results in better performance for both the CCL and MMCL loss functions on both architectures, in our settings.
MMCL consistently delivers the best results on the Yelp and Gowalla datasets.
For Yelp, without early-stopping, MMCL outperforms CCL by 3.98\% in recall and 4.14\% in Ndcg on SimpleX and by 4.31\% in recall and 5.26\% in Ndcg on MF. 
On Gowalla, MMCL gains 7.46\% in recall and 7.05\% in Ndcg on SimpleX, and 1.85\% in recall and 0.63\% in Ndcg on MF.
The relatively lower performance gain on Gowalla is attributed to dataset sparsity, which affects MMCL's ability to learn sufficiently from positive instances while considering all non-trivial negatives.

\textit{\textbf{Training Runtime Comparison}}
We compare the training runtime of the CCL and MMCL loss functions using the MF architecture in Table~\ref{table:compare_efficiency}. The table shows the average time per epoch, change ratio and total training time on the Yelp and Gowalla datasets when with different numbers of negative samples. 
The results indicate that the time per epoch increases as the number of negative samples increases. Training with 800 negative samples takes about 2 hours, while with 100 samples it takes about 1 hour. Comparing CCL and MMCL, using multiple margins has a minor impact, increasing the training time per epoch by no more than 2 seconds on average. This small increase, while utilizing negative samples more efficiently, makes MMCL a promising loss function for RS.

\textit{\textbf{Impact of number of negative samples}}
One of MMCL's main objectives is to utilize negative samples better while keeping batch size and number of negatives limited, reducing the memory requirements during training. 
We compared recent loss functions against MMCL across various numbers of negative samples, as shown in Figure \ref{fig:compare_number_of_negatives}. 
MMCL outperforms CCL with many negative samples, though it is not as effective as BSL. However, with 100 or fewer negative samples, MMCL consistently beats the baselines. 
With 10 negative samples, MMCL outperforms CCL by 19.5\% in recall and 20.13\% in Ndcg on Yelp and by 12.99\% in recall and 15.07\% in Ndcg on Gowalla. 
On the Gowalla dataset, MMCL outperforms both CCL and BSL in recall but lags behind BSL in Ndcg. This discrepancy is due to MMCL's effectiveness in learning from 
non-trivial negative instances, while dataset sparsity makes it challenging for MMCL to gain sufficient insights from positive instances.
As all loss functions achieve stability with 100 negatives, this value was used throughout the experiments.

\textit{\textbf{Performance comparison of MMCL and various loss functions}}
The proposed MMCL loss function was compared with various loss functions using the MF architecture as the base model. Table~\ref{table:compare_loss_functions} lists the comparison results. 
The first group includes traditional loss functions commonly used in RS, while the second group contains recent state-of-the-art loss functions.
The table shows that BSL is a strong baseline, outperforming MMCL in recall on the Yelp dataset and in Ndcg on the Gowalla dataset. 
However, MMCL performs equally well in recall and Ndcg on Yelp and achieves the best results in recall on Gowalla.
MMCL outperforms the best traditional loss function, MSE, by 15.27\% in recall and 15.54\% in Ndcg on Yelp and 30.55\% in recall and 22.54\% in Ndcg on Gowalla.

\section{Conclusion}\label{conclusion}
Recommender systems (RS) guide users by estimating their preferences using methods like collaborative filtering and deep learning.
While complex neural networks are common in RS, simpler models with effective loss functions can perform equally well or better.
Recent trends focus on optimizing loss functions and negative sampling strategies, emphasizing contrastive learning. 
In contrastive learning, it's crucial to effectively utilize negative samples without requiring significant memory resources.
In order to address the challenges of efficient contrastive learning, we propose a new loss function called Multi-Margin Loss (MML). MML uses all non-trivial instances and assigns weights based on their difficulty, enabling effective learning with fewer or smaller batch sizes. Experiments on two real-world datasets showed that MML performs as well as or better than other loss functions, achieving up to a 20\% improvement over baseline contrastive loss functions when fewer number of negative samples are used.

While MML shows notable performance improvements, particularly with smaller negative sample sets, it could benefit from ideas from other loss functions that address issues in RS like popularity bias and diversification. 
Future research could also explore applying MML to other domains beyond recommender systems.




\bibliographystyle{ACM-Reference-Format}
\bibliography{refs}

%

\end{document}